\documentclass[12pt]{report}
\usepackage{setspace}
\usepackage{thesis_style}
\usepackage{coverpage}
\usepackage[utf8]{inputenc}
\usepackage{booktabs, multicol, multirow}
\usepackage{dirtree}
\renewcommand{\sidebarname}{Sidebar}
\usepackage{microtype}
\usepackage{graphicx}
\usepackage{bm}

\usepackage{booktabs} 
\usepackage{makecell}
\usepackage{skak}
\usepackage{amsmath, amsthm, amssymb}
\usepackage[utf8]{inputenc} 
\usepackage[T1]{fontenc}    
\usepackage{hyperref}       
\usepackage{url}            
\usepackage{booktabs}       
\usepackage{amsfonts}       
\usepackage{nicefrac}       
\usepackage{microtype}      
\usepackage{amsmath,amsthm}
\usepackage{graphicx}
\usepackage{caption}
\usepackage{subcaption}

\usepackage{subcaption}

\usepackage{hyperref}
\usepackage{tikz}
\usetikzlibrary{backgrounds}
\usetikzlibrary{calc}
\usetikzlibrary{fit}
\usetikzlibrary{positioning}
\usetikzlibrary{decorations.pathreplacing,calc,shapes,positioning}
\tikzset{>=latex}

\usepackage[shortcuts]{extdash}

\definecolor{mygreen}{HTML}{167dde}
\definecolor{myred}{HTML}{f22835}

\colorlet{greenfill}{mygreen!20!white}
\colorlet{redfill}{myred!20!white}
\colorlet{moreredfill}{myred!40!white}

\newcommand{\E}{\mathbb{E}}

\usepackage{amsmath}
\usepackage{eucal}
\DeclareMathOperator*{\EE}{\mathbb{E}}

\usepackage{amsfonts}

\renewcommand\bar\overline
\renewcommand\epsilon\varepsilon

\usepackage[final]{pdfpages}
\graphicspath{{./images/}}

\usepackage{booktabs}
\usetikzlibrary{trees}
\usepackage{makecell}

\definecolor{mygreen}{HTML}{167dde}
\definecolor{myred}{HTML}{f22835}

\colorlet{greenfill}{mygreen!20!white}
\colorlet{redfill}{myred!20!white}
\colorlet{moreredfill}{myred!40!white}

\usepackage{ifxetex,ifluatex}
\usepackage{etoolbox}

\usepackage{eucal}

\def\1{\bm{1}}

\def\vzero{{\bm{0}}}

\def\vsigma{{\bm{\sigma}}}

\def\mI{{\bm{I}}}

\DeclareMathAlphabet{\mathsfit}{\encodingdefault}{\sfdefault}{m}{sl}
\SetMathAlphabet{\mathsfit}{bold}{\encodingdefault}{\sfdefault}{bx}{n}

\newcommand{\diag}{\mathrm{diag}}

\DeclareMathOperator{\Tr}{Tr}

\newcommand{\q}{{\mathbb Q}}
\newcommand{\p}{{\mathbb P}}

\newcommand{\F}{{\mathcal F}}

\newcommand{\bmx}{{\bm x}}
\newcommand{\bmh}{{\bm h}}

\newcommand{\bmz}{{\bm z}}

\newcolumntype{S}{>{\centering\arraybackslash} m{.08\linewidth} }
\newcolumntype{T}{>{\centering\arraybackslash} m{.10\linewidth} }

\begin{document}

\shorthandoff{:} 

\pagenumbering{roman}

\Auteur{Alex}{Lamb}




\MSc{A Brief Introduction to Generative Models}
    {}
    {of Informatics and Operations Research}
    {University of Montreal}
    {February}
    {2021}

\PagesCouverture

\pagenumbering{arabic}
\setcounter{page}{1}

\onehalfspacing

\chapter{Introduction}
\label{chap:intro}
\section{What are Generative Models?}

One of the most distinctive and powerful aspects of human cognition is the ability to imagine: to synthesize mental objects which are not bound by what is immediately present in reality.  There are many potential reasons why humans evolved this capability.  One is that it allows humans to do planning by imagining how their actions could effect the future.  Another is that by imagining how the future will unfold, humans can test hypotheses about the dynamics of the world, and learn about its properties without explicit supervision.  The subfield of Machine Learning which aims to endow machines with this same essential capacity to imagine and synthesize new entities is referred to as generative modeling.  Along with these lofty ambitions, there are many practical motivations for generative models.  One common argument is that humans may use generative models as a way of doing supervised and reinforcement learning while using less labeled data.  Consider the task of children learning a new language.  While children do receive occasional explicit supervised feedback, i.e. from their parents telling them that they misspelled a word or that they've overgeneralized a word, this feedback is relatively rare.  Reward signals are perhaps just as rare.  Because of this, cognitive scientists are curious about how humans are able to learn with relatively little overfitting.  They refer to this as the ``poverty of the stimulus'' problem.  An appealing hypothesis is that humans use unsupervised generative models to build robust representations of the world and then use those same representations to do supervised and reinforcement learning from small amounts of explicitly labeled data.  Since humans are constantly absorbing perceptual data (sound, vision, touch), humans should have enormous amounts of unlabeled data which can be used to train generative models without overfitting.  We could also imagine that many of these generative models require learning features which are also useful for supervised learning.  For example, one potential generative model would learn to construct the sequence of future visual stimuli conditioned on a window of past visual stimuli $p(X_{t:T} | X_{1:t})$.  A model capable of performing this task well would need to have a strong model for how the world works (i.e. what things form objects, what things are near and far away, what things are large and small, whether something is alive or not, etc.).  

Another practical motivation for generative models is that they could provide better evaluations of the effectiveness of Machine Learning models.  Because classifiers produce the same output for a wide class of inputs, it can be hard to evaluate what a classifier has really learned about a type of data.  Suppose we have a dataset where a model generates text captions from an image.  Consider an image of a giraffe, which the model describes as ``A giraffe walking next to tall green grass''.  It's possible that the model has learned enough to be able to recognize that there is a giraffe, and that the giraffe is walking (and not running or sitting), and that there is tall green grass.  However, another possibility is that the model recognizes the giraffe, but simply says that the giraffe is walking because giraffes are usually walking, and says that the giraffe is near tall grass because giraffes are usually near tall grass, and says that the grass is green because grass is usually green.  Thus it's difficult to know if the model really understands the image, or if it's merely making reasonable guesses based on what types of images are common.  However, consider what could be done if we had a generative model which produced sample images conditioned on the caption $P(X | C)$.  Since humans can generate arbitrary text by hand, we could easily supply the model with counterfactuals like ``A running giraffe next to short red grass'' or ``A giraffe lying down next to tall blue grass''.  Since humans easily generate detailed counterfactuals in text, it would be easy to verify how well the model understands the world.  

\subsection{Formalizing the Generative Modeling Problem}

So far, we have discussed generative modeling in qualitative terms.  We want models which can simulate from the dynamics of the world.  We want models that can synthesize realistic looking data.  However, before going further it is useful to understand the probabilistic interpretation of generative model, which gives a formal framework for studying generative models.  The essential idea is that we treat observations from the world as samples from a distribution $x \sim p(x)$.  For example, we could consider the distribution over all human faces which can occur in reality to be $p(x)$ and consider each face as a sample.  If we have access to a recorded dataset (for example a set of faces), we may also choose to treat these points as a finite collection of samples from this distribution.  

At the same time, we can interpret our generative model as an estimating distribution $q_\theta(x)$, which is described by a set of parameters $\theta$.  Then we can frame generative modeling as trying to ensure that $p(x)$ and $q_\theta(x)$ become as similar as possible.  Statistical divergences give a natural mathematical framework for this.  A divergence is a function $D(p||q) : S \times S \rightarrow R$ taking two distributions $p$ and $q$ over a space of distributions $S$ as inputs, with the properties: 

\begin{align}
D(p||q) \geq 0.0 \\
D(p||q) = 0.0 \iff p = q
\end{align}

Notably, there is no symmetry assumption, so in general $D(p||q) \neq D(q||p)$.  The probabilistic approach to generative modeling frames learning as an optimization problem where the loss corresponds to a given divergence.  

\begin{align}
\mathcal{L}(\theta) = \operatorname*{argmin}_\theta D(p||q_\theta(x))
\end{align}

\subsubsection{Alternatives to the Probabilistic Generative Models Formalization}

At this point, it is worth noting that not all generative models use the probabilistic generative model framework.  Deep style transfer and texture synthesis \citep{gatys2015style,gatys2015texture} search for an image with ``style features'', defined using a fixed neural network, matching a real image.  Deep Image Prior searches for network parameters which produce a single real image (or part of a real image, in the case of inpainting).  Many of these approaches have the distinctive property that they define a rule for modifying a single real image \citep{gatys2015style,ulyanov2017deepprior}, which limits their ability to generalize.  There is also a line of work which has competing objectives: one which encourages the generations to have novel traits and another which encourages the generations to look realistic.  One such example is the Creative Adversarial Network \citep{elgammal2017can} which has one objective encouraging the generated images to differ from styles that occur in the data and another objective encouraging the generated images to follow the data distribution.  These approaches radically differ from the probabilistic framework in that they encourage the production of data points which do not have density under any observed distribution $p(x)$.  

\section{Algorithms}

We briefly overview the two major approaches which are used for probabilistic generative modeling with deep learning.  The first, and considerably older approach, defines a density for generative model and directly maximizes the value of this density on observed data points.  A newer and quite distinctive approach involves modeling the difference between a given generative model and the real data, and then encouraging the generative model to minimize that distance.  

\subsection{The Likelihood Maximization Approach}

What is the right algorithm for finding a distribution $q_\theta(x)$ which minimizes a divergence between itself and $p(x)$.  Before selecting the type of divergence to minimize, a natural question is to consider what types of expressions we are capable of optimizing, and work backwards to find a suitable divergence.  In general, we only have access to samples from the distribution $p(x)$ and not any additional information about the distribution.  At the same time, $q_\theta(x)$ is a model that we control, so it's reasonable to believe that we'll be able to design it so that it has a density that we can compute as well as the ability to draw samples.  


The KL-divergence can be rewritten as an expression in which the only term that depends on the parameters is an expectation on $q_\theta(x)$ over samples from $p(x)$.  Beginning with two distributions $p(x)$ (the empirical distribution) and $q(x)$ (the model distribution), we write the KL-divergence \citep{Nowak-lecture}.

\begin{align}
D_{KL}(p(x) || q(x)) &= \int p(x) \log p(x) dx - \int p(x) \log q(x) dx \\
                     &= \int p(x) \log \frac{p(x)}{q(x)} dx \\
                     &= \mathbb{E}_{x \sim p}[\log \frac{p(x)}{q(x)}] \\
                     &= \mathbb{E}_{x \sim p}[\log p(x) - \log q(x)]
\end{align}

Then we can show the maximum likelihood estimation for a set of $N$ data points.
\begin{align}
\label{equation:mle}
\theta* &= \arg \max_{\theta} \prod_{i=1}^{N} q(x_i) \\
        &= \arg \max_{\theta} \sum_{i=1}^{N} \log q(x_i) \\
        &= \arg \min_{\theta} \frac{1}{N} \sum_{i=1}^{N} -\log q(x_i) \\
        &\sim= \arg \min_{\theta} \mathbb{E}_{l \rightarrow \infty, x \sim p}[-\log q(x)] 
\end{align}

The objective for maximum likelihood is maximizing the log-density $log(q_\theta(x))$ over real data points sampled from the distribution $p(x)$.  

\begin{align}
\label{equation:hxe}
D_{KL}(p(x) || q(x)) &= \int p(x) \log p(x) dx - \int p(x) \log q(x) dx \\
                     &= -H(p(x)) + CE(p(x), q(x)) 
\end{align}

Thus we can see that the KL-divergence decomposes into two terms \ref{equation:hxe}: a cross-entropy term (likelihood) and a term for the entropy of the true data distribution.  Because the entropy of the true data distribution doesn't depend on the estimator, the KL-divergence can be minimized by maximizing likelihood.  Another useful consequence of this is that the entropy of the true data distribution can be estimated by such a generative model if it maximizes likelihood.  

\subsubsection{Estimation with Tractable Densities}

Now that we have established a statistical divergence that we can minimize by optimizing over the parameters of a distribution $\q_\theta$, we turn to the question of deciding what to use as $\q$, which will occupy our attention for the remainder of the section on the maximum likelihood approach.  

The maximum likelihood approach only requires that we be able to sample uniformly from the real data and evaluate the log-density of our estimating distribution $q_\theta(x)$ at these points.  What is the simplest choice for $q$, if we want to frame our problem in terms of optimizing over functions?  Indeed, $q$ cannot simply be an arbitrary function, because it could simply assign a high value to every point in the space.  For $q$ to correspond to the density of an actual probability distribution, it only needs to satisfy two simple properties \ref{equation:posprobdef}: that it be non-negative everywhere and integrate to 1.0 over the region where its value is defined (called the support of the distribution).  To simplify, we'll write the definition using the real numbers $\mathbb{R}$ as the support.  

\begin{align}
\label{equation:posprobdef}
q(x) \geq 0 \\
\label{equation:integralprobdef}
\int_{x \in \mathbb{R}} q(x) = 1
\end{align}

One of the most straightforward ways to satisfy \ref{equation:posprobdef} and \ref{equation:integralprobdef} is to analytically prove that functions with specific parameters satisfy these properties.  While this approach is very limited many such useful functions have been derived and are in widespread use.  One of the most prominent is the normal distribution, which has a density parameterized by $\mu$ and $\sigma$.  

\begin{align}
\label{equation:gaussian}
q(x) &= \frac{1}{\sqrt{2\pi \sigma}} \exp{\frac{-(x - \mu)^2}{2\sigma^2}}
\end{align}

The gaussian integral has a simple value, allowing for a straightforward proof for \ref{equation:integralprobdef} for a variety of distributions involving an exponential over the variable x, which includes the exponential family.  



\subsubsection{Mixture Models}

A major limitation of most closed-form densities, is that they are unimodal, which makes them ill-suited to problems where very distinct points can have high density with regions of low density separating them.  For example, nearly all natural distributions such as image and audio datasets are multimodal.  Moreover, most of these tractable closed-form densities either assume independence between dimensions or only pairwise dependence (for example, the multivariate gaussian).  This further limits the applicability to real data, where density is generally concentrated along specific low-dimensional manifolds \citep{bengio2012mix}.  

One straightforward way to get around these limitations is to replace any density with a mixture over densities with distinct parameters.  This can greatly increase the capacity of the model and has been used in deep generative models \citep{Graves2012,salimans2017pixelcnn++}.  It has a simple closed-form, where each component in the mixture has a weighting $\pi_k$ and a distribution $q_{\theta_{k}}$.  

\begin{align}
\label{equation:mixture}
q_\theta(x) &= \sum_{k=1}^{C} \pi_{k} q_{\theta_{k}}(x)
\end{align}

This form is guaranteed to be normalized with the only condition that the $\pi_k$ sum to 1.0.  This approach has the clear advantage that it allows for a much higher capacity model, yet it has the issue that the only way to achieve more modes is to add more mixture components.  This turns out to be a significant limitation that restricts the utility of mixture models to relatively simple, low-dimensional distributions when the mixture components are unimodal distributions.  

\subsection{Energy-Based Models}

Considering the properties of a distribution in \ref{equation:posprobdef} and \ref{equation:integralprobdef}, one potential path forward is to recognize that the non-negativity is quite easy to enforce by construction, while the constraint on the integral often requires a non-trivial proof.  Based on this, we can define a function called an energy, which is non-negative by construction, but does not necessarily sum to 1.0 over its support.  One way to define this energy is by using an exponential.  

\begin{align}
\label{equation:energy}
E_\theta(x) &= e^{-f_\theta(x)}
\end{align}

We can then compute the sum over the space which x occupies, which we refer to as $Z_\theta$.  

\begin{align}
\label{equation:partition}
Z_\theta    &= \int_{x \in R} e^{-f_\theta(x)} \\\nonumber\\
q_\theta(x) &= \frac{e^{-f_\theta(x)}}{Z_\theta}
\end{align}

When $Z_\theta = 1$, our energy function already defines a probability distribution.  Otherwise, we can divide the energy by $Z_\theta$ to have a probability distribution which satisfies \ref{equation:integralprobdef}.  For energy functions defined by neural networks, there are no general methods of determining this integral over the entire space.  Typically neural networks are evaluated at specific points from some distribution (for example, a data distribution), which is insufficient for computing the integral over the entire space.  

However, computing the gradient of the log-likelihood for energy-based models reveals an interesting form.  

\begin{align}
\label{equation:energy_gradient}
q_\theta(x) &= \frac{e^{-f_\theta(x)}}{Z_\theta}\\
log(q_\theta(x)) &= -f_\theta(x) - log(Z_\theta)\\
-\frac{d log(q_\theta(x))}{d \theta} &= \frac{d f_\theta(x)}{d \theta} + \frac{d log(Z_\theta)}{d \theta}\\
\frac{d log(Z_\theta)}{d \theta} &= \frac{d}{d\theta}\bigg[log\sum_{\Tilde{x}} e^{-f_\theta(x)}\bigg]\\
\frac{d log(Z_\theta)}{d \theta} &= \frac{1}{Z_\theta} \sum_{\Tilde{x}}  \frac{d}{d \theta}\bigg[e^{-f_\theta(\Tilde{x})}\bigg] \\
\frac{d log(Z_\theta)}{d \theta} &= \sum_{\Tilde{x}} -q(\Tilde{x}) \frac{d f_\theta(\Tilde{x})}{d \theta} \\
-\frac{d log(Z_\theta)}{d \theta} &=  \mathop{\mathbb{E}}_{\Tilde{x} \in q_\theta(x)} \bigg[\frac{d f_\theta(\Tilde{x})}{d \theta}\bigg]
\end{align}

When we consider the expected value of the gradient over multiple data samples, the resulting gradient then has a particularly elegant form consisting of a positive phase, in which the function is maximized over samples from the real data, and a negative phase in which the function's value is pushed down at samples from the model distribution.  

\begin{align}
\label{equation:energy_form}
\mathop{\mathbb{E}}_{x \in p(x)} \bigg[-\frac{d log(q_\theta(x))}{d \theta}\bigg] = 
\mathop{\mathbb{E}}_{x \in p(x)} \bigg[\frac{d f_\theta(x)}{d \theta}\bigg] - 
\mathop{\mathbb{E}}_{\Tilde{x} \in q_\theta(x)} \bigg[\frac{d f_\theta(\Tilde{x})}{d \theta}\bigg]
\end{align}

Although this gives a simple form for the gradient, it assumes that we have the ability to sample from our model distribution as well as calculate its energy.  In practice this has been a major obstacle to the use of energy-based probabilistic generative models.  Some research has explored the use of boltzmann machine and an approximation called contrastive divergence which replaces the model distribution $q_\theta(x)$ in \ref{equation:energy_form} with short gibbs sampling chains which start from the real data distribution.  The discovery of a general way of defining energy functions with deep networks which also allow for fast and exact sampling would make energy-based models significantly more appealing.  

Another challenge with energy-based models is that this straightforward form only applies to the gradient of the likelihood with respect to the parameters.  Computing the likelihood itself still turns out to be quite difficult, due to the partition function being an integral over the entire space, while neural networks are typically only evaluated at specific points.  Two solutions have been proposed for estimating the partition function, Annealed Importance Sampling \citep{neal1998ais} and Reverse Annealed Importance Sampling \citep{burda2014raise}, however both are only approximations and require an iterative procedure.  

\subsubsection{Autoregressive Models}

The previous approaches that we've explored for likelihood maximization have tried to increase the expressiveness of $q_\theta(x)$, but this has proved to be difficult for complicated multivariate distributions.  An alternative way to achieve the goal of increased expressiveness for multivariate distributions is to factorize the joint distribution into a chain of conditionals.  Often this can be done with an RNN in the context of sequence modeling.  The density is represented via a fully-observed directed graphical model: it decomposes the distribution over the discrete time sequence $x_{1}, x_{2}, \dots x_{T}$ into an ordered product of conditional distributions over tokens
\[
q_\theta(x_{1}, x_{2}, \dots x_{T}) = q_\theta(x_{1}) \prod_{t=1}^{T} q_\theta(x_{t} \mid x_{1}, \dots x_{t-1}).
\]
When using this autoregressive approach to train RNNs in practice, it is known as \emph{teacher forcing}~\citep{williams1989learning}, due to the use of the ground-truth samples $y_{t}$ being fed back into the model to be conditioned on for the prediction of later outputs (analogous to a teacher directly replacing a student's attempted steps in a solution with correctly completed steps so that they may continue learning). These fed back samples force the RNN to stay close to the ground-truth sequence during training.  

When sampling from an autoregressive model, the ground-truth sequence is not available for conditioning and we sample from the joint distribution over the sequence by sampling each $y_t$ from its conditional distribution given the previously generated samples. Unfortunately, this procedure can result in problems in generation as small prediction error compound in the conditioning context. This can lead to poor prediction performance as the conditioning context (the sequence of previously generated samples) diverges from the distribution seen during training.

~\cite{bengio2015scheduled} proposed to address this exposure bias issue by sometimes feeding the model's predicted values back into the network as inputs during training (as is done during sampling from autoregressive models). However, when the model generates several consecutive $y_t$'s, it is not clear anymore that the correct target (in terms of its distribution) remains the one in the ground truth sequence.  In general these sampled values could be completely inconsistent with the ground truth sequence providing targets for the outputs. This is mitigated in various ways, by making the self-generated subsequences short and annealing the probability of using self-generated vs ground truth samples. However, as remarked by \citet{huszar2015hownot}, scheduled sampling yields a biased estimator, in that even as the number of examples and the capacity go to infinity, this procedure may not converge to the correct model. Nonetheless, in some experiments scheduled sampling still had value as a regularizer.  A consistent way of improving autoregressive models by using adversarial training ~\ref{adv_training} was proposed by \cite{lamb2016professor}.  

In general, the strength of autoregressive models is that they have a straightforward and general statistical formulation in terms of defining a density and directly maximizing likelihood.  Additionally, if each step in the sequence is a scalar, it only requires us to define univariate conditional distributions, and the set of univariate distributions with with closed-form densities is quite general.  For example, a univariate multinomial distribution can be multimodal, can closely approximate a wide range of distributions, and is quite tractable.  

The major weakness of autoregressive models are that the one-step-ahead loss is often not a good fit for long-term measures of error (due to the compounding error effects not observed during training) and that representing uncertainty directly in the space of single steps could be very unnatural.  This may be related to the phenomenon in humans of ``writer's block'', where it's difficult to begin a writing task from scratch.  In the context of autoregressive models, the first few steps often contain a great deal of entropy as they practically constrain the content of all of the text to follow, yet figuring out how the beginning of the text will need to lead to the desired topic or distribution of topics in a long-document can be a challenging task.  Perhaps for this reason, writing often proceeds by an iterative or hierarchical process, instead of as a purely sequential process.  

\subsubsection{Variational Autoencoders}

Another approach for increasing the expressive of learned density functions is to introduce probabilistic latent variables $z$ which capture much of the learned uncertainty, and then represent the distribution $p(x,z) = p(x|z)p(z)$.  Samples from $p(x)$ can then be achieved by marginalizing out over $z$.  

The key appeal of such an approach is that the statistical structure of a learned latent space can often be much simpler than the statistical structure in the visible space.  A density with latent variables has a straightforward form, and from a conceptual perspective, leads to straightforward maximum likelihood estimation.  

\begin{align}
\label{equation:vae_setup}
p(x) &= \prod_{x \in p_{data}(x)} \sum_{z} p(x,z)\\
log(p(x)) &= \sum{x \in p_{data}(x)} \log(\sum_{z} p(x,z))
\end{align}

If the z variable is discrete and has a relatively small number of values, then computing this density is quite straightforward and reasonable.  However, if z is continuous or has many potential values, then computing this sum/integral on each update is either slow or impossible.  It might be tempting to sample a few values of z for each update, and treat the expression as an expectation over both $x$ and $z$.  However this is both biased and quite misguided, as it ignores the interaction between the log and the sum in the expression.  If only a few values of z give rise to a large $p(x,z)$, it's sufficient to give $\log(\sum_{z} p(x,z))$ a large value.  However, if we simply sampled a single $z$ or small number of $z$ on each update, then we would essentially require each $z$ to lead to a $p(x,z)$ with a large value.  Intuitively, it is fine if a few or even a single value of the latent variable explains our observed data, and it is not necessary for all of the $z$ values to explain all of the data points.  

The variational bound provides a mathematical tool for decomposing this likelihood term involving a log-sum structure into a tractable expectation over both $x$ and $z$.  

Introducing the approximate posterior $q_\phi(\mathbf{z} \mid \mathbf{x})$ allows us to decompose the marginal log-likelihood of the data under the generative model in terms of the variational free energy and the Kullback-Leibler divergence between the approximate and true posteriors:

\begin{equation}
\log p_{\bm{\theta}}(\bm{x}) = \mathcal{L}(\theta,\phi;\bm{x}) + D_{\mathrm{KL}}\left(q_{\bm{\phi}}(\bm{z} \mid \bm{x})\|p_{\bm{\theta}}(\bm{z} \mid \bm{x})\right) 
\label{eq:ELBO}
\end{equation}
where the Kullback-Leibler divergence is given by
\begin{equation}
D_{\mathrm{KL}}\left(q_{\bm{\phi}}(\bm{z} \mid \bm{x}) \| p_{\bm{\theta}}(\bm{z} \mid \bm{x})\right) = \E_{q_{\phi}(\bm{z} \mid \bm{x})} \left[ \log \frac{q_{\bm{\phi}}(\bm{z} \mid \bm{x})}{p_{\bm{\theta}}(\bm{z} \mid \bm{x})} \right] \notag
\end{equation}
and the variational free energy is given by
\begin{equation} 
\mathcal{L}(\theta,\phi;\bm{x})= \E_{q_{\bm{\phi}}(\bm{z} \mid \bm{x})}\left[\log \frac{p_{\bm{\theta}}(\bm{x},\bm{z})}{q_{\bm{\phi}}(\bm{z} \mid \bm{x})} \right]. \notag
\end{equation}
Since $D_{\mathrm{KL}}\left(q_{\bm{\phi}}(\bm{z} \mid \bm{x}) \| p_{\bm{\theta}}(\bm{z} \mid \bm{x})\right)$ measures the divergence between $q_{\bm{\phi}}(\bm{z} \mid \bm{x})$ and $p_{\bm{\theta}}(\bm{z} \mid \bm{x})$, it is guaranteed to be non-negative. As a consequence, the variational free energy $\mathcal{L}(\theta,\phi;\bm{x})$ is always a lower bound on the likelihood, which is sometimes referred to as the variational lower bound. 

In the VAE framework, $\mathcal{L}(\theta,\phi;\bm{x})$ is often rearranged into two terms: 
\begin{equation} 
\mathcal{L}(\theta,\phi;\bm{x}) =  \mathcal{L}_{\bm{z}}(\theta,\phi;\bm{x}) + \mathcal{L}_{\bm{x}}(\theta,\phi;\bm{x})
\end{equation}
where
\begin{align}
\mathcal{L}_{\bm{z}}(\theta,\phi;\bm{x}) &= - D_{\mathrm{KL}}\left(q_{\bm{\phi}}(\bm{z} \mid \bm{x}) \| p_{\bm{\theta}}(\bm{z})\right) \notag \\
\mathcal{L}_{\bm{x}}(\theta,\phi;\bm{x}) &= \E_{q_{\bm{\phi}}(\bm{z} \mid \bm{x})}\left[ \log p_{\bm{\theta}}(\bm{x} \mid \bm{z}) \right] \notag 
\end{align}
$\mathcal{L}_{\bm{x}}$ can be interpreted as the (negative) expected reconstruction error of $\bm{x}$ under the conditional likelihood with respect to $q_{\phi}(\bm{z} \mid \bm{x})$.

Maximizing this lower bound pushes the model towards minimizing reconstruction error and minimizing the KL divergence between the approximate posterior $q_{\phi}(\bm{z} \mid \bm{x})$ and the prior $p_{\theta}(\bm{z})$.

With real-valued $\bm{x}$, the \emph{reparametrization trick} 
\cite{kingma2013vae,bengio2013estimating} can be used to
propagate the gradient from the decoder network to the encoder network. The mean of $\bm{x}$ is computed as a deterministic function of
$\bm{x}$ along with the noise term $\epsilon \sim \mathcal{N}(\vzero,\mI)$ such that $\mathbf{z}$ has the desired distribution. Typically the gaussian distribution is used for the posterior.  
\begin{equation}
    q_\phi(\bm{x} \mid \bm{x}) = \mathcal{N}(\bm{x} \mid \mu_\phi(\bm{x}),
                                  \diag(\vsigma^2_\phi(\bm{x}))),
\end{equation}
and the \emph{reparametrization trick} allows the value of a sample to be written in terms of the parameters of the distributed estimated by the network and a noise variable $\epsilon$.  
\begin{equation}
    \bm{x} = \mu_{\phi}(\bm{x}) + \vsigma_{\phi}(\bm{x}) \odot \epsilon,
    \quad \epsilon \sim \mathcal{N}(0, I) \notag
\end{equation}
which produces values with the desired distribution while permitting gradients to propagate into the encoder network through both $\mu_\phi(\bm{x})$ and $\vsigma^2_{\phi}(\bm{x})$. 

\subsubsection{Evaluation Criteria}

Under the maximum likelihood approach, a straightforward way of quantifying the success of the model is to compute the model's average likelihood $q_\theta(x)$ on datapoints from a held-out distribution (often referred to as a test dataset).  This criteria has some desirable qualities: it is able to detect overfitting and it has a consistency guarantee.  On the other hand, it has significant limitations.  One discussed by \citep{arjovsky2017wgan} is that $q_\theta(x)$ has a value approaching 0 at any points where $p(x)$ has a value greater than zero, the log-likelihood approaches negative infinity (intuitively this can be seen by observing that $log(x)$ approaches negative infinity as x approaches 0).  This property is unlikely to be suitable for most applications.  

\subsection{The Adversarial Approach}
\label{adv_training}

An alternative to the likelihood maximization approach involves studying a candidate generative model and the differences between the samples from this model and the original data.  In practice this usually takes the form of estimating the density ratio between the generating distribution $q_\theta(x)$ and the real data distribution $p(x)$.  

\begin{equation}
D_\theta(x) = \frac{p(x)}{q_\theta(x) + p(x)}
\end{equation}

A key motivation behind this approach is that the learning procedure for $D_\theta(x)$ is equivalent to learning a classifier between the real data and the model's distribution.  Classifiers have been extremely successful in deep learning and methods for successfully training classifiers have been widely studied, and inductive biases that are known to be good for classification could also be good for determining the quality of generations.  Another motivation for modeling the difference between a model and the data is that it allows the model to become sensitive to any clear difference between real samples and generated samples, which may be a much easier task than simply determining the density of a distribution at a given point.  

\subsubsection{Noise Contrastive Estimation}

\citep{gutmann2010nce} proposed to use a fixed generative model $q_\theta(x)$ and learn a classifier to distinguish between these samples and the real data distribution.  Once this density ratio $D_\theta(x)$ is learned, the estimator can be sampled from by using importance sampling.  A markov chain monte carlo method could also be used for sampling.  

\begin{align}
D_\theta(x) &= \frac{p(x)}{q_\theta(x) + p(x)}\\
\hat{p}(x) &= \frac{D(x)}{1 - D(x)} q_\theta(x)
\end{align}

A significant limitation in this approach is that a $q_\theta(x)$ must be selected which is very similar to $p(x)$.  For example, the expression isn't even well defined if $q_\theta(x)$ doesn't have support everywhere that $p(x)$ has support.  And if $q_\theta(x)$ has very small values where $p(x)$ has large values, this pushes $D_\theta(x)$ to 1.0, which leads to very large importance weights and high variance sampling.  Intuitively, in a high-dimensional space like an image, a random prior such as a gaussian distribution for $q_\theta(x)$ has no realistic chance of ever producing a realistic image, even though it can happen in theory.  

\subsubsection{Generative Adversarial Networks}

Generative Adversarial Networks (GANs) \citep{goodfellow2014gan}, aimed to leverage the strengths of using a classifier for generation, while avoiding the major weaknesses of noise contrastive estimation.  The GAN framework approachs the generative modeling problem from a perspective inspired by game theory.  The model involves training two networks in an adversarial fashion.  Rather than using a fixed $q_\theta(x)$, a generator network is trained to produce samples which are similar to the training examples and a discriminator network $D_\theta(x)$ is trained to classify between examples from the training set and examples produced by the generator.  The generator is optimized to maximize the probability that the discriminator will classify the generated example as ``real''.  This setup is described as adversarial because the loss for the generator's loss is the opposite of a term in the discriminator's loss.  

\begin{align}
\min_\theta \left [\ell(\q_\theta; \F) := \sup_{F \in \F} F \left( \p, \q_\theta \right) \right]\,.
\label{eq:f-objective}
\end{align}

For the usual cross-entropy classification objective, this can be rewritten more directly.  

\begin{align}
    V(G, D) := \E_{x\sim p_{\rm data}(\bmx)} [ \log D(\bmx)] +  \E_{\bmz\sim q(\bmz)} [\log(1-D(G(\bmz)))], \label{eq:advloss}
\end{align}

A practical observation from \citep{goodfellow2014gan} is that optimizing the generator network to maximize the value of $V(G,D)$ performs poorly in practice, especially when the support of $p$ and $q$ don't overlap.  Instead a non-saturing objective is often used.  

\begin{align}
\max_{\theta} \E_{\bmz\sim q(\bmz)} [\log(D(G_\theta(\bmz)))]
\label{eq:f-objective}
\end{align}

It is also possible to jointly learn inference networks for latent variables $p_{\theta}(\bm{z} | \bm{x})$ in the adversarial framework \citep{dumoulin2017adversarially,lamb2017gibbsnet}.  

\subsubsection{Principled Methods for Training GANs}

\citep{arjovsky2017towards} showed that the gradient for a GAN generator is not well behaved when the support of $q_\theta(x)$ and $p(x)$ don't overlap.  More concretely, they showed that \ref{eq:advloss} leads to saturation and vanishing gradients for the generator and \ref{eq:f-objective} leads to instability and a lack of useful gradient signal in this situation.  

They proposed injecting noise into both $p(x)$ and $q_\theta(x)$ as a way of overcoming this issue in theory, while still learning an estimator of the data distribution.  

\begin{align}
F_\gamma(\p,\q;\phi) := F(\p \ast \Lambda, \q \ast \Lambda; \phi), \quad \Lambda = \mathcal N(\mathbf 0, \gamma \mathbf I)\,.
\label{eq:f-convolved}
\end{align}

In this model, the generator is also trained on gradient from samples with the noise injected.  So long as sufficient noise is injected, this provides the generator with a density ratio which is well-defined even when the support of the real data and generator distribution don't overlap.  It was also shown that the divergence between the distributions with noise injected gives an upper bound on the wasserstein distance between the real distribution and estimating distribution where the tightness of the bound depends on the variance of the noise, which could be controlled by annealing the noise over the course of training.  

\subsubsection{Wasserstein GAN}

A serious problem with GAN training, noted even in its original formulation \citep{goodfellow2014gan} is that on difficult problems, especially early in training, it is difficult to select a generator which has overlapping support with the real data distribution without adding noise (which tends to degrade sample quality).  When the generator and the real data distribution do not have overlapping support, the KL-divergence is undefined and the Jensen-Shannon divergence can be discontinuous around these points.  While \citep{roth2017stable} proposed an analytical approximation to noise injection, the Wasserstein GAN proposes an alternative approach in which a statistical divergence is used which is continuous and differentiable even when the generating distribution and the real data distribution do not overlap.  

A major contribution of \citep{arjovsky2017wgan} is a formulation of a GAN objective which corresponds to optimizing the Earth Mover's distance, or wasserstein metric.  This is based on the Kantorovich-Rubenstein duality which gives a definition of the wasserstein metric in terms of a supremum over all 1-Lipschitz continuous functions $f$.  

\begin{equation} \label{eq:dualitywgan}
W(p, q_\theta) = \sup_{\|f\|_L \leq 1} \EE_{x \sim p}
[f(x)] - \EE_{x \sim q_\theta}[f(x)]
\end{equation}

In practice, this is achieved by using a neural network discriminator as the function $f$.  The lipschitz constraint on f was enforced in \citep{arjovsky2017wgan} by clipping all of the weights to be within a specified range after each update.  \citep{gulrajaniC2017wgangp} proposed to use a penalty on the norm of the gradient of the discriminator's output with respect to its inputs.  This achieved significant improvements over the clipping approach used in the original WGAN.  \citep{roth2017stable} showed that applying the gradient penalty on the original GAN formulation can also achieve good results in practice and can be justified as an analytical approximation to injecting noise into the samples as was theoretically discussed in \citep{arjovsky2017towards}.  

\subsubsection{Spectral Normalization}

\citep{miyato2018spectral} provided a further refinement over gradient penalty based on two primary critiques: (1) that the gradient penalty only guarantees lipschitz continuity at the data points or wherever it is applied, and not everywhere in the input space and (2) that the gradient penalty has the effect of pushing down the rank of the weight matrices and lowering the expressiveness of the discriminator.  \citep{miyato2018spectral} showed that the lipschitz constant of the discriminator function is an upper-bound on the lipschitz constant of the function.  

\begin{align}
    \|f\|_{\rm Lip} \leq & \|(\bmh_{L}\mapsto W^{L+1}\bmh_{L})\|_{\rm Lip}\cdot \|a_L\|_{\rm Lip}\cdot\|(\bmh_{L-1}\mapsto W^ L\bmh_{L-1})\|_{\rm Lip} \nonumber \\
    &\cdots \|a_1\|_{\rm Lip}\cdot\|(\bmh_0\mapsto W^ 1\bmh_0)\|_{\rm Lip}
      = \prod_{l=1}^{L+1} \|(\bmh_{l-1}\mapsto W^l\bmh_{l-1})\|_{\rm Lip} = \prod_{l=1}^{L+1} \sigma(W^l). \label{eq:ineq_lip}
\end{align}

The lipschitz constant $\|g\|_{\rm Lip}$ for a single linear layer $g: \bmh_{in} \mapsto \bmh_{out}$ is equal to the spectral norm of the weight matrix A, which is equivalent to its largest singular value (the largest eigenvalue of $A^*A$).  

\begin{align}
    \sigma(A) := \max_{\bmh: \bmh \neq {\bm 0}} \frac{\|A\bmh\|_2}{\|\bmh\|_2} = \max_{\|\bmh\|_2\leq 1} \|A\bmh\|_2, \label{eq:spnorm}
\end{align}

Therefore, for a linear layer $g(\bmh) = W\bmh$, the norm is given by $\|g\|_{\rm Lip} = \sup_\bmh \sigma(\nabla g(\bmh)) = \sup_\bmh \sigma(W) = \sigma(W) $.

Spectral normalization directly normalizes the spectral norm of the weight matrix $W$ so that it satisfies the Lipschitz constraint $\sigma(W) = 1$:
\begin{align}
\bar{W}_{\rm SN}(W) := W / \sigma(W) \label{eq:sn}.
\end{align}
By normalizing all linear layers in this way, the inequality~\eqref{eq:ineq_lip} and the fact that $\sigma\left(\bar{W}_{\rm SN}(W)\right) = 1$ to see that $\|f\|_{\rm Lip}$ is bounded from above by $1$.

In practice, the eigenvalue of the largest singular value for each weight matrix is maintained by using the power method with a persistent estimate of the eigenvector corresponding to the largest eigenvalue.  The power method consists of an iterative process of multiplying by matrix and re-normalizing.  That the power method results in the eigenvector with the largest eigenvalue may be seen by considering its application on the Jordan-Normal form of the matrix, where the diagonal matrix containing the eigenvalues has the relative value of all but the largest eigenvalue decay with successive iterations.  

As the eigenvector is only a single value and the weights change relatively slowly, the spectral normalization method has almost no computational cost, unlike the gradient penalty.  

Aside from its computational advantages, there are two major advantages to spectral normalization over the gradient penalty.  The first is that spectral normalization only penalizes based on the size of the largest eigenvalue, so there is no pressure to reduce the rank of the weight matrices.  Moreover, dividing a matrix by a non-zero constant does not change its rank.  On the other hand, \citep{miyato2018spectral} showed that weight normalization and gradient penalty both have the effect of pushing down the rank of the weight matrices.  This could make it more difficult for the network to learn expressive functions.  A second advantage of spectral normalization is that it enforces the lipschitz constraint at all points in the space, whereas gradient penalty only enforces the lipschitz constraint at points where it is applied, usually around data points or on linear interpolations between data points.  

\subsubsection{Jacobian Clamping}

So far we have looked at approaches for improving GAN training which consider modifications to the discriminator and its training objective.  An alternative and potentially complementary approach was explored in \citep{odena2018jacobian} which consists of an additional objective which encourages the generator's output to not change too much or too little as the latent value $z$ is changed by a small amount.  The eigenvalues $\lambda_1,\lambda_2,...$ and corresponding eigenvectors $v_1, v_2, ...$ of the metric tensor associated with $G(z)$ and its jacobian can be written as follows \citep{odena2018jacobian}.  

\begin{equation}
  \label{eqn:eqlimitjacobian}
\lim_{||\epsilon|| \to 0} \frac{||G(z) - G(z + \epsilon v_k)||}{
  ||\epsilon v_k||} = \sqrt{\lambda_k}
\end{equation}

The condition number is given by the ratio $\frac{\lambda_k}{\lambda_1}$.  The metric tensor is considered to be poorly conditioned if the condition number has a high value.  \citep{odena2018jacobian} proposed to eschew the issue of computing the complete spectrum, which could be quite challenging, in favor of sampling random directions (essentially sampling small random values for $\epsilon v_k$ and empirically computing ~\ref{eqn:eqlimitjacobian}, and then adding a penalty to encourage these values to fall within a specific range.  In practice they achieved good results by setting $\lambda_{min}$ to 1.0 and $\lambda_{max}$ to 20.0.  Making $\lambda_{max}$ too small could have the effect of making it too hard for the model to be responsive to the latent variables and setting $\lambda_{min}$ to be too large could have the effect of making it impossible for the model to learn to give large regions in the space relatively constant density.  In practice these hyperparameters would likely need to be tuned depending on the dataset in accordance with these concerns.  

\subsubsection{Evaluation Criteria}

The maximum likelihood approach provided a straightforward, if not completely well motivated, way to quantitatively evaluate generative models within its family.  For adversarial approaches, no such criteria is readily apparent.  Why is this?  The discriminator's score provides an estimate of how much the model's density differs from the true data density at a given point.  If the discriminator is able to correctly classify between real data points and generated data points reliably and in a way that generalizes, then it is a clear indicator that the generator is of poor quality.  However, if the opposite is true, that the discriminator cannot classify between real and fake, then it could either be because the generator is of high quality, or it could be because the discriminator is somehow limited (in architecture, training procedure, or another characteristic).  This means that the discriminator's score cannot reliably used as a way of discerning the quality of a generative model (although in the case of the Wasserstein GAN, it is at least informative enough to gauge the progress of training) \citep{arjovsky2017wgan,gulrajaniC2017wgangp}.  

This basic limitation has motivated the exploration of novel quantitative evaluation criteria for generative models in the adversarial family.  Despite having this motivation for their development, both of the criteria that we will discuss are agnostic to the actual form of the generative model, and could equally be applied to models trained using maximum likelihood.  

Two different but very closely related methods have seen widespread adoption as methods for quantitatively evaluating adversarially trained generative models.  In both cases a fixed pre-trained classifier is used as the basis for the scoring metric.  The first is the Inception Score \citep{salimans2016improved}, which is defined by:

\begin{equation}
\exp\left (\mathbb{E}_{\mathbf{x} \in q_\theta} [KL(p(y \vert \mathbf{x}) \Vert p(y)]  \right)
\end{equation}

where $\mathbf{x}$ is a GAN sample, $p(y\vert \mathbf{x})$ is the probability for labels $y$ given by a pre-trained classifier on $\mathbf{x}$, and $p(y)$ is the marginal distribution of the labels in the generated samples according to the classifier.  Higher scores are considered better.  The intuition behind this metric is that a generator should produce samples from many different classes while at the same time ensuring that each sample is clearly identifiable as belonging from a single class.  For example, a generator which only produces samples of a single class will have a poor inception score because $p(y)$ and $p(y \vert x)$ will be very similar, as it will only reflect that single class.  Likewise producing samples which do not give the classifier clear information about the class will tend to make $p(y|x)$ uncertain and more similar to $p(y)$, leading to a poor inception score.  

While inception score has been shown to be highly correlated to visual sample quality (real data samples achieve better inception scores than any current models) and tends to give bad scores to clearly deficient GAN models (as well as models early in training), three limitations in Inception Score are readily apparent.  One is that the inception score could be pushed beyond the values achievable with real data by a model which produces only a single and clearly identifiable example of each class that the classifier is aware of.  This would make $p(y \vert x)$ very different from $p(y)$ while having high entropy in $p(y)$, and yet the model would clearly lack the diversity of real data, and would be a poor generative model from the statistical divergence perspective.  Another limitation is that if the classifier is vulnerable to adversarial examples, this could hypothetically be exploited to achieve unnaturally high inception scores.  This was demonstrated directly in experiments by \citep{barratt2018inception}.  While this is potentially an issue if researchers are unscrupulous and in a competitive setting, it is unclear if this will occur if a researcher does not intentionally set out to produce adversarial examples for the inception score classifier.  Finally, a straightforward problem with inception score is that a very high score can be achieved just by returning the samples from the training set.  Thus a generative model which merely memorizes the training data would achieve a high inception score, without doing any learning.  The inception score will give low scores to model which underfits and fails to achieve clear samples, but it does not penalize a model at all for memorizing the training data or failing to generalize.  

The Frechet Inception Distance \citep{heusel2017frechet} was proposed to address some of the limitations inherent in the inception score.  It shares the idea of using a fixed pre-trained classifier as its foundation, but instead of assessing the quality of $p(y \vert x)$ for samples, it instead takes hidden activations from the end of the classifier.  The key idea is that the score is high when the distribution of these activations for generated samples is close to the distribution of these activations for real data points.  How can we determine if these distributions are indeed close to each other, without simply reproducing the problem of having to train a generative model?  While this remains an open question, the proposal in FID is to assume that these hidden states follow a multivariate gaussian distribution (but not necessarily with an isotropic variance).  Because these hidden states are from the end of a deep classifier, this multivariate gaussian assumption is much more justified than it would be in the visible space.  

To compute the FID score, one fits a multivariate gaussian $N(m, C)$ to the activations for the real test samples and fits a separate multivariate gaussian $N(m_w, C_w)$ to the activations for the generated samples.  From this, the Frechet Distance between the distributions has a surprisingly tractable form which does not require inverting the covariance matrices $C$ or $C_w$.  

\begin{equation}
\|m-m_w\|_2^2+ \Tr \bigl(C+C_w-2\bigl(CC_w\bigr)^{1/2}\big)
\end{equation}

\citep{heusel2017frechet} studied several artificial deformations of the data and showed that FID scores gradually became worse with increasing corruption.  More specifically: they studied artifically injecting independent noise into the images, removing random regions of the images, swirling the images, salt and pepper noise, and injecting examples from another dataset.  On all of these increasing corruption led to worse FID scores, whereas only injecting unrelated samples led to worse inception scores.  

Perhaps most important, FID can be evaluated on the test data, so it can directly test against overfitting, unlike inception scores.  Moreover, generating a single high quality example for each class (at the expense of overall diversity) could still hurt FID by giving the hidden states of the generated samples an unnatural distribution.  While these metrics are now widely used in measuring the quality of generative models \citep{odena2018jacobian,miyato2018spectral,karras2017progressive}, they are still highly dependent on the choice of the classifier for evaluation and lack statistical consistency guarantees.  

\cleardoublepage\pagebreak
\bibliographystyle{chicago}
\markboth{Bibliography}{Bibliography}
\addcontentsline{toc}{chapter}{\numberline{}Bibliography}
\bibliography{strings,ml,aigaion,junyoung,myrefs,lamb}

\end{document}